\documentclass[pmlr]{jmlr}

\usepackage{longtable}

 \usepackage{booktabs}
 \usepackage{csquotes}

 \usepackage{svg}
 \svgsetup{inkscapelatex=false}
 
\usepackage[load-configurations=version-1]{siunitx} 

\makeatletter
\def\set@curr@file#1{\def\@curr@file{#1}} 
\makeatother

\theorembodyfont{\upshape}
\theoremheaderfont{\scshape}
\theorempostheader{:}
\theoremsep{\newline}

\jmlrvolume{}
\jmlryear{2023}
\jmlrworkshop{}

\usepackage{booktabs, multirow, makecell} 
\usepackage{soul}
\usepackage{changepage,threeparttable} 

\title[CancerGPT]{CancerGPT: Few-shot Drug Pair Synergy Prediction using Large Pre-trained Language Models}

\author{\Name{Tianhao Li*}
       \Email{tianhao@utexas.edu}\\ 
       \addr School of Information\\
       University of Texas at Austin\\
       Austin, Texas, USA
       \AND
       \Name{Sandesh Shetty*}
       \Email{sandeshshett@umass.edu}\\ 
       \addr College of Information and Computer Sciences\\
       University of Massachusetts Amherst\\
       Amherst, MA, USA
       \AND
       \Name{Advaith Kamath}
       \Email{advaith.kamath@utexas.edu}\\ 
       \addr Department of Chemical Engineering\\
       University of Texas at Austin\\
       Austin, TX, USA 
       \AND
       \Name{Ajay Jaiswal}
       \Email{ajayjaiswal@utexas.edu}\\ 
       \addr School of Information\\
       University of Texas at Austin\\
       Austin, TX, USA
       \AND
        \Name{Xiaoqian Jiang}
       \Email{xiaoqian.jiang@uth.tmc.edu}\\ 
       \addr School of Biomedical Informatics\\
       University of Texas Health Science Center at Houston\\
       Houston, TX, USA
       \AND
       \Name{Ying Ding}
       \Email{ying.ding@ischool.utexas.edu}\\ 
       \addr School of Information\\
       University of Texas at Austin\\
       Austin, TX, USA
       \AND
        \Name{Yejin Kim}
       \Email{yejin.kim@uth.tmc.edu}\\ 
       \addr School of Biomedical Informatics\\
       University of Texas Health Science Center at Houston\\
       Houston, TX, USA} 

\usepackage{lineno}

\begin{document}

\maketitle

\begin{abstract}
Large pre-trained language models (LLMs) have been shown to have significant potential in few-shot learning across various fields, even with minimal training data. However, their ability to generalize to unseen tasks in more complex fields, such as biology, has yet to be fully evaluated. LLMs can offer a promising alternative approach for biological inference, particularly in cases where structured data and sample size are limited, by extracting prior knowledge from text corpora. Our proposed few-shot learning approach uses LLMs to predict the synergy of drug pairs in rare tissues that lack structured data and features. Our experiments, which involved seven rare tissues from different cancer types, demonstrated that the LLM-based prediction model achieved significant accuracy with very few or zero samples. Our proposed model, the CancerGPT (with $\sim$ 124M parameters), was even comparable to the larger fine-tuned GPT-3 model (with $\sim$ 175B parameters).
Our research is the first to tackle drug pair synergy prediction in rare tissues with limited data. We are also the first to utilize an LLM-based prediction model for biological reaction prediction tasks.

\end{abstract}

\section{Introduction}


Foundation models have become the latest generation of artificial intelligence (AI) (\cite{Moor2023-dp}). Instead of designing AI models that solve specific tasks one at a time, such foundation models or ``generalist'' model can be applied to many downstream tasks without specific training. For example,  large pre-trained language model (LLM), such as GPT-3 (\cite{brown_language_2020}) and GPT-4 (\cite{OpenAI2023-ce}), has been a game changer in foundation AI model (\cite{Mitchell2023-cs}). LLM can apply its skills to unfamiliar tasks that it has never been trained for, which is few-shot learning or zero-shot learning. This is due in part to multitask learning, which enables LLM to unintentionally gain knowledge from implicit tasks in its training corpus (\cite{radford_language_nodate}). Although LLM has shown its proficiency in few-shot learning in various fields (\cite{brown_language_2020}), including natural language processing, robotics, and computer vision (\cite{veit_learning_2017, brown_language_2020, wertheimer_few-shot_2019}), the generalizability of LLM to unseen tasks in more complex fields such as biology has yet to be fully tested. In order to infer unseen biological reactions, knowledge of participating entities (e.g., genes, cells) and underlying biological mechanisms (e.g., pathways, genetic background, cellular environment) is required. While structured databases encode only a small portion of this knowledge, the vast majority is stored in free-text literature which could be used to train LLMs. Thus, we envision that, when there are limited structured data and limited sample sizes, LLMs can serve as an innovative approach for biological prediction tasks, by extracting prior knowledge from unstructured literature. One of such few-shot biological prediction tasks with a pressing need is a drug pair synergy prediction in understudied cancer types.  

\begin{figure}[t]
  \centering 
  \includegraphics[width=5in]{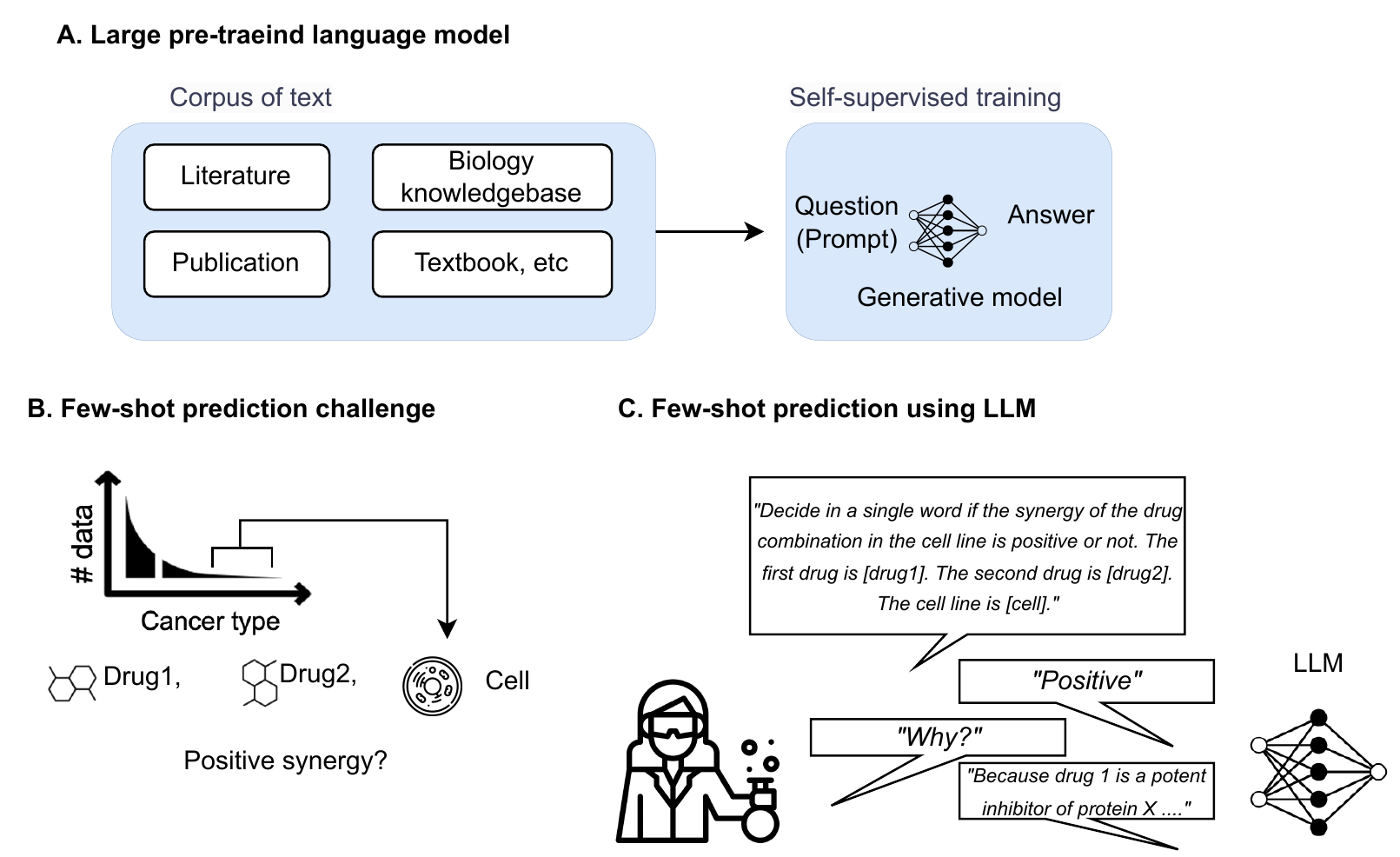}
  \caption{Few-shot prediction in biology. A. Different from task-specific approach, large pre-trained language model can perform new tasks which are not been explicitly trained for. B. Drug pair synergy prediction in rare tissues is an important examples of numerous few-shot prediction tasks in biology. C. Large pre-trained language model can be an innovative approach for few-shot prediction in biology thanks to its prior knowledge encoded in its weight.}
  \label{fig:motivation} 
\end{figure}

Drug combination therapy has become a widely accepted strategy for treating complex diseases such as cancer, infectious diseases, and neurological disorders. In many cases, combination therapy can provide better treatment outcomes than single-drug therapy. Predicting drug pair synergy has become an important area of research in drug discovery and development. Drug pair synergy refers to the enhancement of the therapeutic effects of two (or more) drugs when used together compared to when each drug is used alone. The prediction of drug pair synergy can be challenging due to a large number of possible combinations and the complexity of the underlying biological mechanisms (\cite{zagidullin_drugcomb_2019}). Several computational methods have been developed to predict drug pair synergy, particularly using machine learning. Machine learning algorithms can be trained on large datasets of in vitro experiment results of drug pairs to identify patterns and predict the likelihood of synergy for a new drug pair. However, most of the data available comes from common cancer types in certain tissues, such as breast and lung cancer; very limited experiment data are available on certain types of tissues, such as bone and soft tissues (Fig. \ref{fig:motivation}). Obtaining cell lines from these tissues can be physically difficult and expensive, which limits the number of training data available for drug pair synergy prediction. This can make it challenging to train machine learning models that rely on large datasets. 

Early studies in this area have relied on relational information or contextual information to extrapolate the synergy score to cell lines in other tissues, (\cite{chen_drugcom_2018, 10.1093/bioinformatics/btaa287, li_network_2018, kuru_matchmaker_2022, 10.1093/bioinformatics/btac579}), ignoring the biological and cellular differences in these tissues. Another line of studies has sought to overcome the discrepancy between tissues by utilizing diverse and high-dimensional features, including genomic (e.g., gene expression of cell lines) or chemical profiles (e.g., drug structure) (\cite{preuer_deepsynergy_2018, liu_transynergy_2021, kuru_matchmaker_2022, hosseini_ccsynergy_2023, kim_anticancer_2021}). Despite the promising results in some tissues (with abundant data), these approaches cannot be applied to tissues with too limited data to adapt its model with the large number of parameters for those high-dimensional features. 

In this work, we aim to overcome the above challenge by LLMs. We hypothesize that cancer types with limited structured data and discrepant features still have good information in scientific literature.  Manually extracting predictive information on such biological entities from literature is a complex task. Our innovative approach is to leverage prior knowledge in scientific literature encoded in LLMs. We built a few-shot drug pair synergy prediction model that transforms the prediction task into a natural language inference task and generate answers based on prior knowledge encoded in LLMs. Our experimental results demonstrate that our LLM-based few-shot prediction model achieved significant accuracy even in zero shot setting (i.e., no training data) and outperformed strong tabular prediction models in most cases. 

This remarkable few-shot prediction performance in one of the most challenging biological prediction tasks has a critical and timely implication to a broad community of biomedicine because it shows a strong promise in the ``generalist'' biomedical artificial intelligence (\cite{Moor2023-dp}). 

\section{Related Work}
\label{sec:related_works}

\subsection{Drug pair synergy prediction}
Lots of methods have been proposed to predict drug pair synergy in recent years. Based on the data type to use, these methods can be classified either as a multi-way relational method or as a context-aware method. Multi-way relational methods (\cite{chen_drugcom_2018, 10.1093/bioinformatics/btaa287, 10.1093/bioinformatics/btac579}) use drug and cell line's relational information without any further chemical or gene information as input and predict drug pair’s synergy. Context-aware methods (\cite{preuer_deepsynergy_2018, liu_transynergy_2021, kuru_matchmaker_2022, hosseini_ccsynergy_2023}) further utilized chemical and gene information from drugs and cell lines to predict drug pair's synergy, which usually contains drug-drug, drug-gene, gene-gene interactions, and cellular environment. These methods usually achieve good performance with rich features on common tissues. However, both approaches do not apply to the cell lines in rare tissues with the limited size of data and cellular information. \cite{kim_anticancer_2021} uses transfer learning to extend the prediction model trained in common tissues to some of the rare tissues with relatively rich data and cellular features. However, it cannot be utilized for rare tissues with extremely limited data and cellular information.

\subsection{Few-shot learning on tabular data}
Traditional supervised learning algorithms can struggle due to the difficulty in obtaining enough labeled data for classification. Few-shot learning is an emerging field that aims to address this issue by enabling machines to learn from a few examples rather than requiring a large size of labeled data. Meta-learning (\cite{finn_model-agnostic_2017, Wang2023-iw, Gao2023-se}) is one technique for few-shot learning. It trains a model on a set of tasks in a way that allows it to quickly learn to solve new, unseen tasks with a few examples. Another technique is data augmentation (\cite{nam_stunt_2023, Yang2022-ne}), which generates new examples by transforming existing data. One promising but less explored direction is to leverage LLMs, particularly when prior knowledge encoded in a corpus of text can be served as a predictive feature. TabLLM (\cite{hegselmann_tabllm_2023}) is one such framework.  It serializes the tabular input into a natural language text and prompts LLM to generate predictions. Leveraging TabLLM, we investigated the effectiveness of LLMs in few-shot learning tasks in biology.

\subsection{Language models for biomolecular sequence analysis}
There has been a growing interest in using language models for biomolecular sequence analysis, and one approach involves the training of language models with biomolecular data (\cite{Madani2023-sp, nvidia_bionemo}). These models learn the language of biomolecules, such as DNA, RNA, and protein sequences, similar to how GPT-2 (\cite{radford_language_nodate}) or GPT-3 (\cite{brown_language_2020}) learns human language. However, our study takes a different approach. Rather than training a language model specifically for biomolecular data, we use a  language model that has been pre-trained on a corpus of human language text. This pre-trained model is used as a few-shot prediction model for drug pair synergy data, allowing us to make accurate predictions with minimal training data.  By leveraging the power of pre-trained language models, we are able to make use of existing resources and obtain generalizability to diverse biological prediction tasks beyond biomolecule sequence analysis.  

\section{Results}

\begin{figure}[t]
  \centering 
  \includegraphics[width=7in]{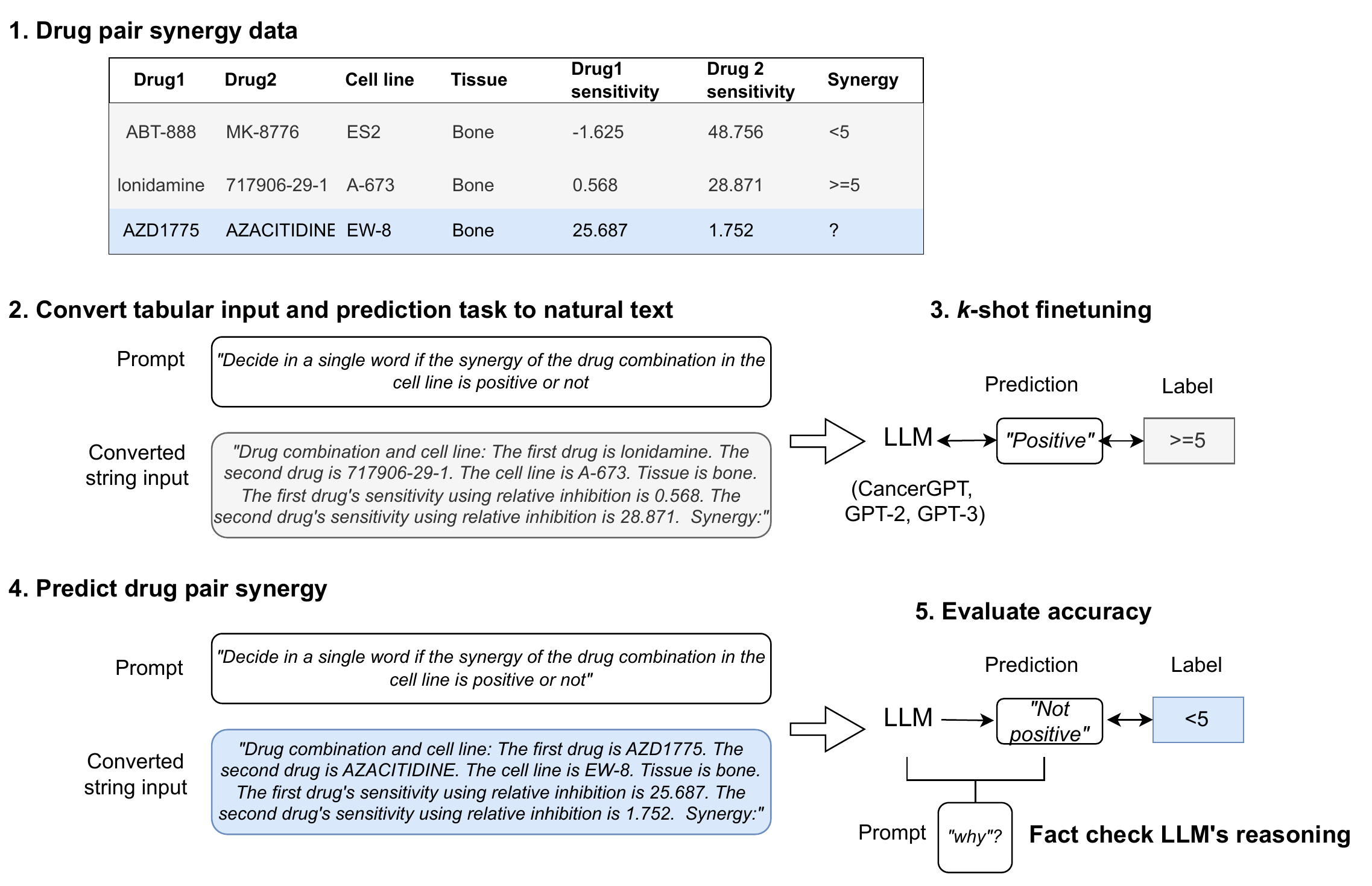}
  \caption{Study workflow. We first converted the tabular input to natural text and created a task-specific prompt (Section \ref{sec:convert}). The prompt was designed to generate binary class predictions (e.g., \emph{``Positive'', ``Not positive''}). We fine-tuned the LLMs (GPT-2 and GPT-3) with $k$-shots of data in rare tissues (Section \ref{sec:k_shot}). We further tailored GPT-2 by fine-tuning it with a large amount of common tissue data, in order to adjust GPT-2 in the context of drug pair synergy prediction (CancerGPT, Section \ref{sec:cancergpt2}). We evaluated and compared the prediction models with a different number of shots and tissues (Section \ref{sec:experiments}). We investigated the LLM's reasoning based on factual evidence.}  
  \label{fig:study_workflow} 
\end{figure}

We developed CancerGPT, a few-shot drug pair synergy prediction model for rare tissues. Leveraging LLMs-based tabular data prediction model (\cite{hegselmann_tabllm_2023}), we first converted the prediction task into a natural language inference task and generated answer using prior knowledge from the scientific literature encoded in LLM's pre-trained weight matrices (Section \ref{sec:llm-prediction-model}, Fig. \ref{fig:study_workflow}). We presented our strategy to adapt the LLM to our task with only a few shots of training data in each rare tissue in Section \ref{sec:k_shot} and Fig. \ref{fig:models}.

To evaluate the performance of our proposed CancerGPT model and other LLM-based models, we conduct a series of experiments in which we compare the model with various other tabular models (Section \ref{sec:experiments}). We measured accuracy using the area under the precision-recall curve (AUPRC) and the area under the receiver operating curve (AUROC) under the different settings. We considered different few-shot learning scenarios, where the model is provided with a limited number $k$ of training data to learn from ($k$=0 to 128). By varying the number of shots, we can examine the model's ability to adapt and generalize with minimal training data. 
Next, we investigated the performance of CancerGPT and other LLM-based models across different tissue types. Since cancer is a highly heterogeneous disease with various subtypes, it is crucial for the model to be able to accurately predict outcomes in diverse tissue contexts. We then investigated whether the LLM's reasoning for its prediction is valid by checking its argument with scientific literature. 

\subsection{Accuracy}

\begin{adjustwidth}{-2.5cm}{-2.5cm}\centering\begin{threeparttable}[!htb]
\scriptsize
\begin{tabular}{lrrrrrrrrrr}\toprule
& &\multicolumn{8}{c}{Number of shots} \\\cmidrule{3-10}
&Methods &0 &2 &4 &8 &16 &32 &64 &128 \\\midrule
\multirowcell{5}{Pancreas \\ ($n_0$=38, $n_1$=1)} &XGBoost &0.026 &- &- &- &- &- &- &- \\
&TabTransformer &0.056 &- &- &- &- &- &- &- \\
&CancerGPT &0.033 &- &- &- &- &- &- &- \\
&GPT-2 &0.032 &- &- &- &- &- &- &- \\
&GPT-3 &\textbf{0.111} &- &- &- &- &- &- &- \\
& & & & & & & & & \\
\multirowcell{5}{Endometrium \\ ($n_0$=36, $n_1$=32)} &XGBoost &0.5 &0.5 &0.5 &0.5 &0.5 &0.5 &- &- \\
&TabTransformer &0.674 &0.889 &0.903 &\textbf{0.948} &\textbf{0.938} &\textbf{0.962} &- &- \\
&CancerGPT &0.564 &0.668 &0.676 &0.831 &0.686 &0.737 &- &- \\
&GPT-2 &0.408 &0.808 &0.395 &0.383 &0.389 &0.717 &- &- \\
&GPT-3 &\textbf{0.869} &\textbf{1} &\textbf{0.947} &0.859 &0.799 &0.859 &- &- \\ \\
\multirowcell{5}{Liver \\ ($n_0$=192, $n_1$=21)} &XGBoost &0.132 &0.132 &0.132 &0.132 &0.132 &0.132 &0.12 &0.12 \\
&TabTransformer &0.13 &\textbf{0.128} &0.147 &0.189 &0.265 &0.168 &0.169 &0.234 \\
&CancerGPT &0.136 &0.102 &0.13 &0.147 &0.252 &0.21 &0.197 &0.187 \\
&GPT-2 &\textbf{0.5} &0.099 &\textbf{0.151} &\textbf{0.383} &\textbf{0.429} &\textbf{0.401} &\textbf{0.483} &0.398 \\
&GPT-3 &0.185 &0.086 &0.096 &0.125 &0.124 &0.314 &0.362 &\textbf{0.519} \\ \\
\multirowcell{5}{Soft tissue \\ ($n_0$=269, $n_1$=83)} &XGBoost &0.243 &0.243 &0.243 &0.243 &0.235 &0.235 &0.264 &0.271 \\
&TabTransformer &0.273 &0.287 &\textbf{0.462} &\textbf{0.422} &\textbf{0.526} &0.571 &0.561 &0.64 \\
&CancerGPT &\textbf{0.314} &\textbf{0.315} &0.338 &0.383 &0.383 &0.403 &0.464 &0.469 \\
&GPT-2 &0.259 &0.298 &0.254 &0.262 &0.235 &0.297 &0.254 &0.206 \\
&GPT-3 &0.263 &0.194 &0.28 &0.228 &0.363 &\textbf{0.618} &\textbf{0.638} &\textbf{0.734} \\ \\
\multirowcell{5}{Stomach \\ ($n_0$=1081, $n_1$=109)} &XGBoost &0.104 &0.104 &0.104 &0.104 &0.104 &0.104 &0.09 &0.094 \\
&TabTransformer &0.261 &\textbf{0.371} &\textbf{0.396} &\textbf{0.383} &\textbf{0.294} &\textbf{0.402} &\textbf{0.45} &\textbf{0.465} \\
&CancerGPT &\textbf{0.3} &0.297 &0.316 &0.325 &0.269 &0.308 &0.297 &0.312 \\
&GPT-2 &0.116 &0.124 &0.099 &0.172 &0.165 &0.107 &0.152 &0.131 \\
&GPT-3 &0.078 &0.106 &0.17 &0.37 &0.1 &0.19 &0.219 &0.181 \\ \\
\multirowcell{5}{Urinary tract \\ ($n_0$=1996, $n_1$=462)} &XGBoost &0.186 &0.186 &0.186 &0.186 &0.186 &0.197 &0.199 &0.209 \\
&TabTransformer &0.248 &\textbf{0.264} &\textbf{0.25} &\textbf{0.278} &\textbf{0.274} &0.249 &\textbf{0.293} &\textbf{0.291} \\
&CancerGPT &0.241 &0.226 &0.246 &0.239 &0.256 &\textbf{0.271} &0.266 &0.269 \\
&GPT-2 &0.191 &0.192 &0.188 &0.156 &0.193 &0.185 &0.183 &0.185 \\
&GPT-3 &\textbf{0.27} &0.228 &0.222 &0.201 &0.206 &0.2 &0.24 &0.272 \\ \\
\multirowcell{5}{Bone \\ ($n_0$=3732, $n_1$=253)} &XGBoost &0.064 &0.064 &0.064 &0.064 &0.064 &0.064 &0.064 &0.064 \\
&TabTransformer &\textbf{0.123} &0.12 &0.121 &0.115 &0.102 &\textbf{0.13} &\textbf{0.129} &0.121 \\
&CancerGPT &0.119 &\textbf{0.115} &\textbf{0.125} &\textbf{0.116} &\textbf{0.115} &0.11 &0.114 &0.125 \\
&GPT-2 &0.063 &0.094 &0.057 &0.081 &0.052 &0.071 &0.057 &0.065 \\
&GPT-3 &0.064 &0.051 &0.045 &0.058 &0.068 &0.087 &0.101 &\textbf{0.181} \\
\bottomrule
\end{tabular}
\caption{AUPRC of $k$-shot learning on seven tissue sets. $n_0$:=total number of non-synergistic samples (not positive), $n_1$:=total number of synergistic samples (positive). We used 20\% data as a test set in each rare tissue, while ensuring the binary labels were equally represented.}
\label{tab:auprc}
\end{threeparttable}\end{adjustwidth}
\begin{adjustwidth}{-2.5cm}{-2.5cm}\centering\begin{threeparttable}[!htb]
\scriptsize
\begin{tabular}{lrrrrrrrrrr}\toprule
& &\multicolumn{8}{c}{Number of shots} \\\cmidrule{3-10}
&Methods &0 &2 &4 &8 &16 &32 &64 &128 \\\midrule
\multirowcell{5}{Pancreas } &XGBoost &0.5 & & & & &- &- &- \\
&TabTransformer &0.553 & & & & &- &- &- \\
&CancerGPT &0.237 & & & & & & & \\
&GPT-2 &0.211 & & & & &- &- &- \\
&GPT-3 &\textbf{0.789} & & & & & & & \\
& & & & & & & & & \\
\multirowcell{5}{Endometrium} &XGBoost &0.5 &0.5 &0.5 &0.5 &0.5 &0.5 &- &- \\
&TabTransformer &0.694 &0.857 &0.878 &\textbf{0.939} &\textbf{0.939} &\textbf{0.959} &- &- \\
&CancerGPT &0.489 &0.693 &0.714 &0.735 &0.612 &0.612 &- &- \\
&GPT-2 &0.265 &0.816 &0.224 &0.184 &0.204 &0.612 &- &- \\
&GPT-3 &\textbf{0.837} &\textbf{1} &\textbf{0.949} &0.898 &0.878 &0.898 &- &- \\
& & & & & & & & & \\
\multirowcell{5}{Liver} &XGBoost &0.587 &0.587 &0.587 &0.587 &0.587 &0.587 &0.574 &0.574 \\
&TabTransformer &0.535 &0.506 &0.526 &0.535 &0.609 &0.647 &0.702 &0.804 \\
&CancerGPT &0.615 &0.468 &0.59 &0.641 &\textbf{0.782} &\textbf{0.776} &\textbf{0.737} &0.737 \\
&GPT-2 &\textbf{0.731} &0.449 &0.558 &\textbf{0.66} &0.679 &0.763 &0.731 &0.731 \\
&GPT-3 &0.615 &0.49 &0.542 &0.583 &0.474 &0.731 &\textbf{0.737} &\textbf{0.91} \\
& & & & & & & & & \\
\multirowcell{5}{Soft tissue} &XGBoost &0.491 &0.491 &0.491 &0.491 &0.454 &0.476 &0.542 &0.552 \\
&TabTransformer &0.557 &0.566 &\textbf{0.709} &0.727 &\textbf{0.788} &\textbf{0.802} &0.83 &0.835 \\
&CancerGPT &\textbf{0.656} &0.646 &0.68 &\textbf{0.734} &0.725 &0.754 &0.8 &0.795 \\
&GPT-2 &0.546 &0.535 &0.519 &0.56 &0.427 &0.577 &0.456 &0.384 \\
&GPT-3 &0.517 &0.406 &0.6 &0.444 &0.607 &0.82 &\textbf{0.866} &\textbf{0.889} \\
& & & & & & & & & \\
\multirowcell{5}{Stomach} &XGBoost &0.529 &0.529 &0.529 &0.529 &0.529 &0.529 &0.476 &0.508 \\
&TabTransformer &\textbf{0.804} &\textbf{0.863} &\textbf{0.855} &\textbf{0.853} &\textbf{0.812} &\textbf{0.85} &\textbf{0.885} &\textbf{0.869} \\
&CancerGPT &0.794 &0.792 &0.796 &0.794 &0.785 &0.787 &0.824 &0.808 \\
&GPT-2 &0.551 &0.569 &0.521 &0.516 &0.589 &0.538 &0.469 &0.566 \\
&GPT-3 &0.419 &0.575 &0.724 &0.769 &0.534 &0.69 &0.742 &0.724 \\
& & & & & & & & & \\
\multirowcell{5}{Urinary tract} &XGBoost &0.494 &0.494 &0.494 &0.494 &0.494 &0.526 &0.53 &0.544 \\
&TabTransformer &0.599 &0.612 &0.604 &0.625 &0.601 &0.587 &0.623 &0.622 \\
&CancerGPT &0.578 &0.561 &0.579 &0.577 &0.589 &0.569 &0.593 &0.609 \\
&GPT-2 &0.526 &0.528 &0.532 &0.397 &0.515 &0.452 &0.469 &0.566 \\
&GPT-3 &\textbf{0.645} &0.57 &0.556 &0.496 &0.508 &0.516 &0.531 &0.572 \\
& & & & & & & & & \\
\multirowcell{5}{Bone} &XGBoost &0.499 &0.499 &0.499 &0.499 &0.499 &0.499 &0.499 &0.499 \\
&TabTransformer &\textbf{0.706} &\textbf{0.705} &\textbf{0.724} &\textbf{0.697} &\textbf{0.65} &\textbf{0.689} &\textbf{0.708} &0.696 \\
&CancerGPT &0.625 &0.648 &0.693 &0.653 &0.636 &0.658 &0.681 &0.68 \\
&GPT-2 &0.507 &0.616 &0.471 &0.579 &0.421 &0.552 &0.476 &0.518 \\
&GPT-3 &0.498 &0.415 &0.341 &0.429 &0.485 &0.605 &0.62 &\textbf{0.794} \\
\bottomrule
\end{tabular}
\caption{AUROC of $k$-shot learning on seven tissues sets.}
\label{tab:auroc}
\end{threeparttable}\end{adjustwidth}

We evaluated the accuracy of our synergy prediction models. We calculated the AUPRC and AUROC of the LLM-based models (CancerGPT, GPT-2, GPT-3) and baseline models (XGBoost, TabTransformer) (Table \ref{tab:auprc}, \ref{tab:auroc}). Due to an imbalance in positive and non-positive labels, we reported both AUPRC and AUROC. Details on the classification task and threshold of synergy are discussed in Section \ref{sec:hyperparameter}. 

\paragraph{Number of training data and accuracy}

Overall, the LLM-based models (CancerGPT, GPT-2, GPT-3) achieved comparable or better accuracy in most of the cases compared to baselines. In the zero-shot scenario, the LLM-based models generally had higher accuracy than the baseline models in all experiments except stomach and bone. As the number of shots increased, we observed mixed patterns across various tissues and models. TabTransformer consistently exhibited an increase in accuracy with more shots. CancerGPT showed higher accuracy with more shots in the endometrium and soft tissue, and GPT-3 showed higher accuracy with more shots in the liver, soft tissues, and bone, indicating that the information gained from a few shots of data complements the prior knowledge encoded in CancerGPT and GPT-3. 

However, the LLM-based models sometimes did not show significant improvements in accuracy in certain tissues, such as the stomach and urinary tract, suggesting that the additional training data do not always improve the LLM-based models' performance. 
With the maximum number of shots ($k$=128), the LLM-based model, specifically GPT-3, was on par with TabTransformer, achieving the highest accuracy with the pancreas, liver, soft tissue, and bone, while TabTransformer achieved the best accuracy with endometrium, stomach, and urinary tract.

\paragraph{Tissue types and accuracy}
The accuracy of the models varied depending on the tissue types, as each tissue possessed unique characteristics and had different data size. In pancreas and endometrium tissues, GPT-3 showed high accuracy with only a few shots ($k$=0 or 2). Generally, the cell lines from the two tissues are difficult to obtain and have a limited number of well-established cell lines, which makes them less investigated. For example, the pancreas is located deep within the abdomen, making it difficult to access and isolate cells without damaging them. The endometrium is a complex tissue that undergoes cyclic changes during the menstrual cycle, and this dynamic process complicates the cell culturing process. Due to this limited training data, few-shot drug pair synergy prediction in these tissues required even higher generalizability.

In the liver, soft tissue, and bone, GPT-3 again achieved the highest accuracy than any other models, including one that trained with common tissues (TabTransformer, CancerGPT). This may be because these tissues have unique cellular characteristics specific to their tissue of origin that training with common tissues may not help predict accurately. For example, hepatic cell lines (originated from liver tissue) are often used in research on drug metabolism and toxicity and have unique drug response characteristics due to high expression of drug-metabolizing enzymes such as cytochrome P450s (\cite{guo_similarities_2011}). Bone cell lines have bone-specific signaling pathways that can affect drug responses, and the extracellular matrix composition and structure in bone tissue can also impact drug delivery and efficacy (\cite{lin_bone_2020}). 

On the other hand, models trained with common tissues (TabTransformer, CancerGPT) achieved the best accuracy in the stomach and urinary tract tissues of all $k$, indicating that the prediction learned from common tissues can be extrapolated to these tissues. Particularly, CancerGPT achieved the highest accuracy with no training sample ($k$=0) in the stomach.  

\paragraph{Comparing LLM-based models}
When comparing LLM-based models, CancerGPT and GPT-3 demonstrated superior accuracy compared to GPT-2 in most tissues. GPT-3 exhibited higher accuracy than CancerGPT in tissues with limited data or unique characteristics, while CancerGPT performed better than GPT-3 in tissues with less distinctive characteristics, such as the stomach and urinary tract. The higher accuracy of CancerGPT compared to GPT-2 highlights that well-balanced adjustment to specific tasks can increase the accuracy while maintaining generalizability. However, the benefits of such adjustments may diminish with larger LLM models, such as GPT-3 (175B parameters), in situations where more generalizability is required.  The fact that CancerGPT with smaller parameters (124M parameters) achieved the comparable accuracy to GPT-3 with larger parameters (175B parameters) implies that further fine-tuning of GPT-3 could achieve even higher accuracy.

\subsection{Fact check LLM's reasoning}
We evaluated whether the LLM can provide the biological reasoning behind its prediction. In this experiment, we used zero-shot GPT-3 because other fine-tuned LLM-based models compromised its language generative performance during the fine-tuning and were not able to provide coherent responses. To do this, we randomly selected one true positive prediction and examined whether its biological rationale was based on factual evidence or mere hallucination. Our example was the drug pair AZD4877 and AZD1208 at cell line T24 for urinary tract tissue. We prompted the LLMs with \emph{``Could you provide details why are the drug1 and drug2 synergistic in the cell line for a given cancer type?’’}. Details on prompt generation are discussed in Supplementary 1.  We evaluated the generated answer by comparing it with existing scientific literature. We found that the LLM provided mostly accurate arguments, except for two cases (Table \ref{tab:reasoning}) in which no scientific literature exists. By combining these individual scientific facts the LLM inferred the unseen synergistic effect. Generally, drugs targeting non-overlapping proteins in similar pathways are more likely to be synergistic (\cite{Cheng2019-dp, Tang2022-wn}).  In  this case, both AZD4877 and AZD1208 target similar pathways that inhibit tumor cell divisions without overlapping protein targets. The Loewe synergy score of this pair at T24 was 46.82, meaning strong positive synergistic effect.

\begin{table}[!htp]
\begin{tabular}{|p{3in}|p{3in}|}\toprule
Excerpt of the generated answer & Fact check and reference \\ \hline
\emph{``The combination of AZD-4877 and AZD1208 has been studied in T24 cells...to be synergistic in reducing bladder cancer cell growth and metastasis''} &False. No study conducted on this drug pair \\ \hline
\emph{``The combination was also found to target multiple pathways involved in the growth and spread of bladder cancer cells.''} &True. AZD1208 is a PIM1 inhibitor. PIM1 is overexpressed in bladder cancer initiation and progression (\cite{Guo2010-ly}). AZD4877 is a drug designed to target bladder cancer (\cite{Jones2013-zl}). \\ \hline
\emph{``...Specifically, AZD-4877 was found to inhibit the activation of proteins involved in the promotion of tumor growth...''} &True. AZD4877 is a drug designed to target bladder cancer (\cite{Jones2013-zl}). \\ \hline
\emph{``...AZD1208 was found to inhibit proteins associated with the inhibition of tumor growth.''} &True. AZD1208 inhibits the cell growth by suppressing p70S6K, 4EBP1 phosphorylation, and messenger RNA translation (in acute myeloid leukemia) (\cite{Cortes2018-tx}). \\ \hline
\emph{``This combination was also effective at reducing the production of inflammatory mediators such as cytokines, which are known to contribute to tumor progression.''} &False. AZD1208 is a pan-PIM kinase inhibitor, and PIM kinases are downstream effectors of cytokine (\cite{noauthor_2011-op}). However, AZD4877 has no evidence in reducing inflammatory mediators. \\ \hline
\emph{``...these two drugs have been shown to reduce levels of apoptosis inhibitors, which can also play a role in tumor progression.''} &True. AZD1208 induce cell apoptosis (\cite{Cervantes-Gomez2019-de}). AZD4877 is a inhibitor of Eg5, which promotes cell apoptosis (\cite{Borthakur2009-wg}). \\
\bottomrule
\end{tabular}
\caption{Example of generated answer when the LLM was asked to provide its reasoning for its prediction}\label{tab:reasoning}
\end{table}

\subsection{Example of prediction results}
As an example, we listed predicted synergistic drug pairs for stomach and soft tissue using CancerCPT (Table S3.1, S3.2) and bone and liver tissue using GPT-3 (Table S3.3, S3.4). We randomly selected two true positive, false positive, true negative, and false negative prediction examples.
We discovered that Loewe synergy scores of the true negative or false negative prediction examples were close to the threshold we used to categorize the label (i.e., Loewe score $>$5). This suggests that accuracy may vary significantly by different thresholds for determining positive synergy. Setting more extreme thresholds (e.g., $>$10, $>$30), like previous models (\cite{kim_anticancer_2021}, \cite{kuru_matchmaker_2022}, \cite{hosseini_ccsynergy_2023}), may increase the prediction accuracy. 


\section{Discussion} 

\paragraph{Summary}
Our study investigates the potential of LLMs as a widely applicable few-shot prediction model in the field of biology. Specifically, we propose a new few-shot model for predicting drug pair synergy, which can be used in rare tissues with few or no training samples available. We transformed tabular data prediction into natural language inference tasks and fine-tuned LLMs (GPT-2, GPT-3) with very few samples in each tissue. The CancerGPT model, which was further tuned with a large amount of common tissue data, showed comparable accuracy to the few-shot tuned GPT-3 model, indicating that tailoring GPT-3 to specific tasks could further improve prediction accuracy. The LLM's reasoning for its prediction revealed that it implicitly infers unseen synergistic effects by combining several independent scientific facts. 

\paragraph{Why drug pair synergy prediction to evaluate LLMs}
The prediction of drug pair synergy in uncommon tissues serves as an excellent benchmark task for evaluating LLMs in few-shot learning within the field of biology. This prediction requires incorporating multiple pieces of information, such as drug and cell line, as well as the sensitivity of drugs to the cell lines, in order to infer the synergistic effects. While detailed information on these entities can be found in scientific papers, the interaction effect, or synergistic effect, is primarily available through biological experiments. To effectively assess LLMs' inference capabilities, one must employ a prediction task where the ground truth is not explicitly available in text format but can be determined through alternative sources for model evaluation. Typically, drug pair synergy scores are obtained through high-throughput testing facilities involving robot arms (\cite{he_methods_2018}). Therefore, individual records of the experiments are seldom recorded in academic literature, decreasing the likelihood of their use as training data for LLMs. Additionally, few studies have been conducted on rare tissues regarding their synergy prediction models, and their synergy prediction outcomes are not explicitly stated in text format. Another similar task is predicting the sensitivity of a single drug in a cell line; however, since the sensitivity of individual drugs is extensively researched and well-documented in publications, the LLM model may merely recollect from the text rather than infer unseen tasks.

\paragraph{Comparison to existing drug pair synergy prediction models}
It should be noted that it was not possible to compare our LLM-based models with previous predictions of drug pair synergy. The majority of models necessitate high-dimensional features of drugs and cells (e.g., genomic or chemical profiles), along with a substantial amount of training data, even the one specifically designed for rare tissue (\cite{kim_anticancer_2021}). This kind of data is not easily accessible in rare tissues, which makes it challenging to carry out a significant comparison. Our model is designed to address a common but often overlooked situation where we have limited features and data. Thus, we compared the LLM-based models with other tabular models that share the same set of inputs.

\paragraph{Contribution}
The contribution of our study can be summarized as follows. In the area of drug pair synergy prediction in rare tissues, our study is the first to predict drug pair synergy on tissues with very limited data and features, which other previous prediction models have neglected. This breakthrough in drug pair synergy prediction could have significant implications for drug development in these cancer types. By accurately predicting which drug pair will have a synergistic effect on these tissues in which cell lines are expensive to obtain, biologists can directly zoom into the most probable drug pairs and perform in vitro experiments in a cost effective manner. 

Our study also delivers generalizable insights about LLMs in the broader context of  biology. To the best of our knowledge, our study was the first to investigate the use of LLMs as an few-shot inference tool based on prior knowledge in the field of biology, where much of the latest information is presented in unstructured free text (such as scientific literature). This innovative approach could have significant implications for advancing computational biology where obtaining abundant training data is not readily possible. By leveraging the vast amounts of unstructured data available in the field, LLMs can help researchers bypassing the challenge of limited training data when building data-driven computational models. 

Furthermore, this LLM-based few-shot prediction approach could be applied to a wide range of diseases beyond cancer, which is currently limited by the scarcity of available data. For instance, this approach could be used in infectious diseases, where the prompt identification of new treatments and diagnostic tools is crucial. LLMs could help researchers quickly identify potential drug targets and biomarkers for these diseases, resulting in faster and more effective treatment development.

\paragraph{Limitations}
The present study, while aiming to showcase the potential of LLMs as a few-shot prediction model in the field of biology, is not without its limitations. To fully establish the generalizability of LLMs as a ``generalist'' artificial intelligence, a wider range of biological prediction tasks must be undertaken to validate it. Additionally, it is crucial to investigate how the information gleaned from LLMs complements the existing genomic or chemical features that have traditionally been the primary source of predictive information. In future research, we plan to delve deeper into this aspect and develop an ensemble method that effectively utilizes both existing structured features and new prior knowledge encoded in LLMs. 

Furthermore, while we observed that GPT-3's reasoning was similar to our own when fact-checking its argument with scientific literature in one example, it is important to note that the accuracy of its arguments cannot always be verified and may be susceptible to hallucination. It is reported that LLMs can also contain biases that humans have (\cite{Schramowski2022-yg}). Therefore, further research is necessary to ensure that the LLM's reasoning is grounded in factual evidence. Despite these limitations, our study provides valuable insights into the potential of LLMs as a few-shot prediction model in biology and lays the groundwork for future research in this area.

\section{Method}

\subsection{Problem Formulation}

\paragraph{Objective}
Our objective is to predict whether a drug pair in a certain cell line has a synergistic effect, particularly focusing on rare tissues with limited training samples. Given an input $$x = \{d_1, d_2, c, t, ri_1, ri_2 \}$$ of drug pair $(d_1, d_2)$, cell line $c$, tissue $t$, and the sensitivity of the two drugs using relative inhibition, the prediction model is $$y \approx f(x) $$ where $y$ is the binary synergy class (1 if positive synergy; 0 otherwise).  Prior research (\cite{10.1093/bioinformatics/btac579, hosseini_ccsynergy_2023}) has employed three different scenarios for predicting drug pair synergy (random split, stratified by cell lines, stratified by drug combinations). Our task is to predict synergy when the data are stratified by tissue, which is a subset of cell lines.

\paragraph{Why tabular input}
As discussed in Section \ref{sec:related_works}, relationships learned in a tissue cannot be well generalizable to other tissues that have different cellular environments. This biological difference poses a challenge in predicting drug pair’s synergy in tissues with a limited number of samples. The limited sample size makes it even more difficult to incorporate typical cell line features, such as gene expression level, which has large dimensionality (e.g., $\sim$ 20,000 genes). Due to this data challenge, the drug pair synergy prediction model is then reduced to build a prediction model with limited samples (few or zero-shot learning) with only limited tabular input feature types. Specific input features were described in Section \ref{sec:experiments}.

\subsection{Synergy prediction models based on Large pre-trained language models}
\label{sec:convert}

\paragraph{Converting tabular input to natural text}
To use an LLM for tabular data, the tabular input and prediction task must be transformed into a natural text. For each instance of tabular data (Fig. \ref{fig:study_workflow}), we converted the structured features into text. For example, given the feature string (e.g., ``drug1'', ``drug 2'', ``cell line'', ``tissue'', ``sensitivity1'', ``sensitivity2'') and its value (e.g., ``lonidamine'', ``717906-29-1'', ``A-673'', ``bone'', ``0.568'', ``28.871''), we converted the instance as \emph{``The first drug is AZD1775. The second drug is AZACITIDINE. The cell line is SF-295. Tissue is bone. The first drug's sensitivity using relative inhibition is 0.568. The second drug's sensitivity using relative inhibition is 28.871.''} Other alternative ways to convert the tabular instance into the natural text are discussed in previous papers (\cite{li_deep_2020, narayan_can_2022}).

\paragraph{Converting prediction task into natural text}
We created a prompt that specifies our tasks and guides the LLM to generate a label of our interest. We experimented with multiple prompts. One example of the prompts we created was \emph{``Determine cancer drug combination synergy for the following drugs. Allowed synergies: {{Positive, Not positive}}. {{ Tabular Input }}. Synergy:''}. As our task is a binary classification, we created the prompt to only generate binary answers (\emph{``Positive'', ``Not positive''}).  Comparing these multiple prompts (Supplementary 1), the final prompt we used in this work was \emph{``Decide in a single word if the synergy of the drug combination in the cell line is positive or not. \{\{ Tabular Input \}\}. Synergy:''.}

\subsection{LLM-based prediction model}
\label{sec:llm-prediction-model}
\paragraph{Large pre-trained language models}

We built our prediction models by tuning GPT-2 and GPT-3 into our tasks (Fig. \ref{fig:study_workflow}).  GPT-2 is a Transformer-based large language model which was pre-trained on a very large corpus of English data without human supervision. It achieved state-of-the-art results on several language modeling datasets in a zero-shot setting when it was released, and it is the predecessor of GPT-3 and GPT-4. GPT-2 (\cite{radford_language_nodate}) has several versions with different sizes of parameters, GPT-2, GPT-Medium, GPT-Large, and GPT-XL. We used GPT-2 with the smallest number of parameters (regular GPT-2, 124 million) in this work to make the model trainable on our server. To adjust the model for a binary classification task, we added a linear layer as a sequence classification head on top of GPT-2, which uses the last token of the output of GPT-2 to classify the input. The cross-entropy loss was used to optimize the model during the fine-tuning process (discussed below). 

 GPT-3 (\cite{brown_language_2020}) is a Transformer-based autoregressive language model with 175 billion parameters, which achieved state-of-the-art performance on many zero-shot and few-shot tasks when it was released. GPT-3.5, including ChatGPT (\cite{openai_chatgpt}), a famous fine-tuned model from GPT-3.5, is an improved version of GPT-3. However, the GPT-3 model and its parameters are not publicly available.  Although the weight of the GPT-3 model is undisclosed, OpenAI offers an API (\cite{openai_finetuning}) to fine-tune the model and evaluate its performance. We utilized this API to build drug pair synergy prediction models through $k$-shot fine-tuning. There are four models provided by OpenAI for fine-tuning, Davinci, Curie, Babbage, and Ada, of which Ada is the fastest model and has comparable performance with larger models for classification tasks. For that reason, we use GPT-3 Ada as our classification model. After uploading the train data, the API adjusted the learning rate, which is 0.05, 0.1, or 0.2 multiplied by the original learning rate based on the size of the data, and fine-tuned the model for four epochs. A model of the last epoch was provided for further evaluation.

 \subsection{CancerGPT}
 \label{sec:cancergpt2}
 We further tailored GPT-2 by fine-tuning it with a large amount of common tissue data, in order to adjust GPT-2 in the context of drug pair synergy prediction. We named this model CancerGPT. CancerGPT used the same structure as the modified GPT-2 mentioned above. A linear layer was added to the top of GPT-2, which uses the last token of the GPT-2 output to predict the label. To use the pre-trained GPT-2 model, the same tokenizer was used as GPT-2. Left padding was used to ensure the last token was from the prompt sentence. The cross-entropy loss was used to optimize the model.  

CancerGPT was first fine-tuned to learn the relational information between drug pairs from common tissues, similar to collaborative filtering (\cite{10.1093/bioinformatics/bty452}) (Fig. \ref{fig:models}). This approach was based on the assumption that certain drug pairs exhibit synergy regardless of the cellular context, and therefore, the relational information between drug pairs in common tissues can be used to predict synergy in new cell lines in different tissues (\cite{hosseini_ccsynergy_2023}). Additionally, we incorporated information on the sensitivity of each individual drug to the given cell line, using relative inhibition score as a measure of sensitivity (\cite{10.1093/nar/gkab438}). By doing so, we were able to gather a more detailed and nuanced understanding of the relationship between drugs and cell lines.

Subsequently, we utilized CancerGPT as one of the pre-trained LLMs and fine-tuned to $k$ shots of data in each rare tissue (as discussed in the following section). All the LLM models use the tabular input that was converted to natural text and share the same prompt.

\subsection{$k$-shot fine-tuning strategy}
\label{sec:k_shot}

\begin{figure}[t]
  \centering 
  \includegraphics[width=6in]{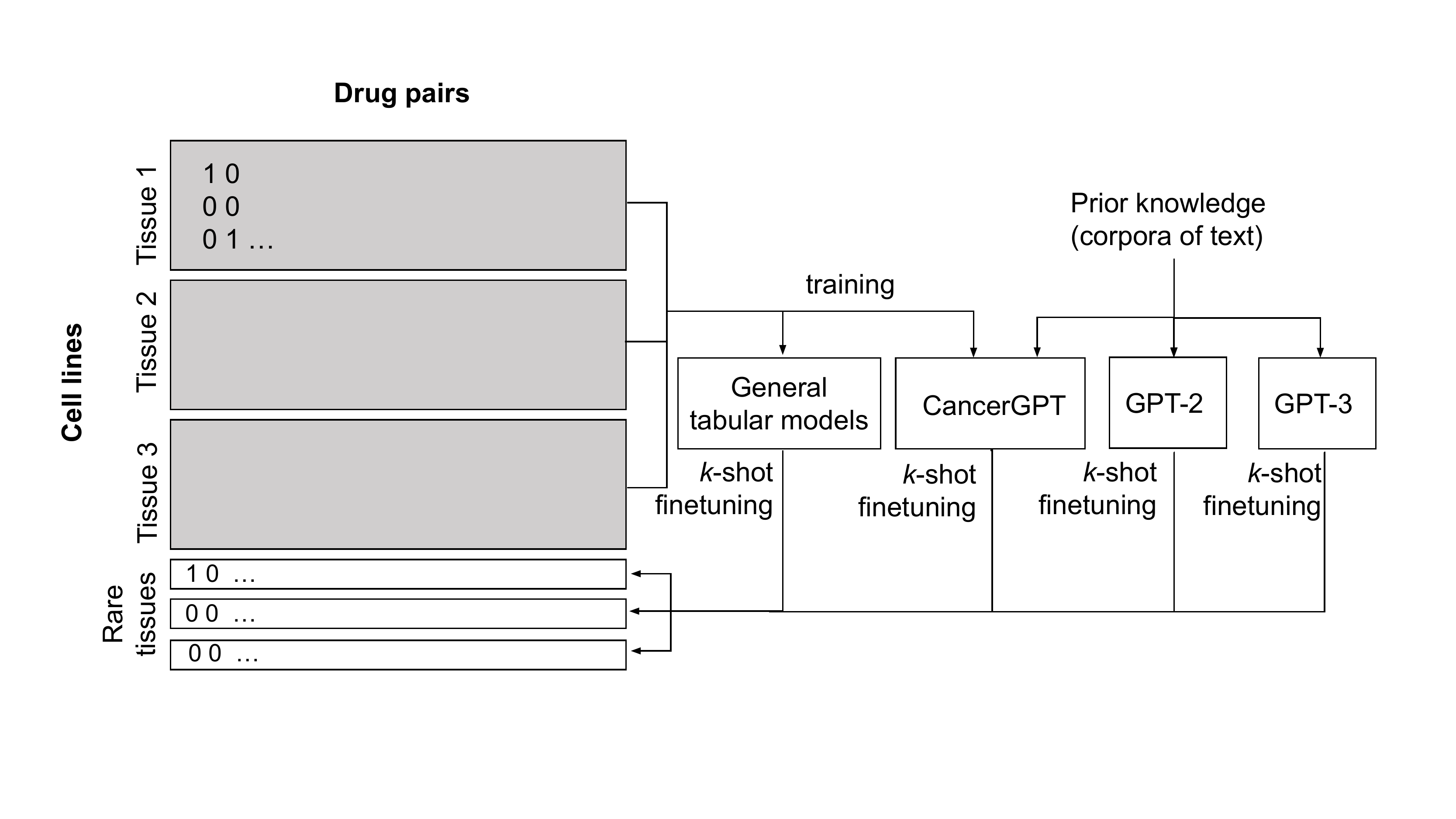}
  \caption{Training strategy of baseline and proposed LLM-based models. General tabular models and CancerGPT were first trained with samples from common tissues then $k$-shot fine-tuned with each tissue of interest. GPT-2 and GPT-3 are pre-trained models, and we fine-tuned them with $k$ shots of data in each tissue.}
  \label{fig:models} 
\end{figure}

The LLM-based models had different training and fine-tuning strategies (Fig. \ref{fig:models}). Samples of common tissues were split into 80\% train data and 20\% validation data for CancerGPT. The models were trained using train data and evaluated by validation data to determine the models with specific hyperparameters to be used for further fine-tuning on rare tissues. For the GPT-2 and GPT-3 based prediction models, we directly used pre-trained parameters from GPT-2 (\cite{radford_language_nodate}) using Huggingface’s Transformers library (\cite{wolf2020huggingfaces}) and GPT-3 Ada from OpenAI (\cite{brown_language_2020}) respectively.

All these models were then fine-tuned with $k$ shots of data in each of the rare tissues. For bone, urinary tract, stomach, soft tissues, and liver, we performed experiments with $k$ from $[0, 2, 4, 8, 16, 32, 64, 128]$. For endometrium and pancreas, because of the limited number of data, we implemented experiments with $k$ from $[0, 2, 4, 8, 16, 32]$ from the endometrium, and only zero shot ($k$ = 0) for the pancreas.

With the limited number of shots, a careful balance of binary labels in the train and test set was critical. We partitioned the data into 80\% for training and 20\% for testing in each rare tissue, while ensuring the binary labels were equally represented in both sets. We randomly selected $k$ shots from the training for fine-tuning, while maintaining consistency with previously selected shots and adding new ones. Specifically, we maintained the previously selected $k$ shots in the training set and incremented additional $k$ shots to create $2 \times k$ shots. The binary label distribution in each $k$ shot set followed that of the original data, with at least one positive and one negative sample included in each set. For evaluation stability, the test data was consistent across different shots for each tissue.

\section{Experiments}
\label{sec:experiments}

\subsection{Dataset}
We utilized a publicly accessible extensive database of drug synergy from DrugComb Portal (\cite{zagidullin_drugcomb_2019}), which is an open-access data portal where the results of drug combination screening studies for a large variety of cancer cell lines are accumulated, standardized, and harmonized. The database contains both drug sensitivity rows and drug pair synergy rows. After filtering the available drug pair synergy rows, the data contains 4,226 unique drugs, 288 cell lines, with a total of 718,002 drug pair synergy rows. We employed the Loewe synergy score, which ranges from -100 (antagonistic effect) to 75 (strong synergistic effect), for drug combination synergy. (\cite{greco_search_1995})  The Loewe synergy score quantifies the excess over the expected response if the two drugs are the same compound (\cite{ianevski_synergyfinder_2017, yadav_searching_2015}). In this paper, we focused on cell lines from rare tissues. We defined the rare tissues as the ones with less than 4000 samples, which include the pancreas (n=39), endometrium (n=68), liver (n=213), soft tissues (n=352), stomach (n=1,190), urinary tract (n=2,458), and bone (n=3985). We tested our models with each of the rare tissues. 

\subsection{Baseline models}
We compared the LLM-based prediction model with two other tabular models that take the same set of inputs. We specifically used XGBoost (\cite{Chen2016-bv}) and TabTransformer (\cite{huang_tabtransformer_2020}). XGBoost is one of the gradient-boosting algorithms for supervised learning based on tree ensemble. for structured or tabular data. It is  widely used in large-scale drug synergy data (\cite{Sidorov2019-np, Celebi2019-nl}).

TabTransformer is a self-attention-based supervised learning model for tabular data. TabTransformer applies a sequence of multi-head attention-based Transformer layers on parametric embeddings to transform them into contextual embeddings, in which highly correlated features will be close to each other in the embedding space. Considering the highly correlated nature of drugs in our data, TabTransformer can be a very strong baseline in this work

To train the two baseline models, we first converted the drugs and cell lines in the tabular data into indicators using one-hot coding. Tissue information was not used in training because the models will be tested in one specific rare tissue that is not used in training. Neither XGBoost nor TabTransformer is a pre-trained LLM; thus, no further contextual information can be inferred through the unseen tissue indicator. For XGBoost, all the variables (drugs, cell lines, and sensitivities) were used as input to predict the drug pair synergy. For TabTransformer, we first trained an embedding layer from scratch on the categorical variables (drugs and cell lines) and passed them through stacked multi-headed attention layers, which we then combined with the continuous variables (sensitivities). This combination then passes through feed-forward layers, which have a classification head.

\subsection{Hyperparameter Setting}
\label{sec:hyperparameter}

The predicted output was a binary label indicating the presence of a synergistic effect, with a Loewe score greater than 5 indicating a positive result. We used AUROC and AUPRC to evaluate the accuracy of classification. Regression tasks were not possible in our LLM-based models because our model can only generate text-based answers (\emph{``positive''} or \emph{``not positive''}), with poor precision in accurately quantifying the synergy value. 

XGBoost was used with a boosting learning rate of 0.3. The number of the gradient boost trees was set to 1000 with a maximum tree depth of 20 for base learners. TabTransformer was used with a learning rate of 0.0001 and a weight decay of 0.01. The model was trained for 50 epochs on common tissues. During the training, the model with the best validation performance was selected for further fine-tuned on rare tissues. For each $k$ shot in each tissue, the model was fine-tuned using the same learning rate and weight decay for 1 epoch and tested with AUPRC and AUROC. Details in the hyperparameter setting are discussed in Supplementary 2.

CancerGPT was first fine-tuned with pre-trained regular GPT-2 for 4 epochs on common tissues. The learning rate was set to be 5e-5 and weight decay was set to be 0.01. Then the model was fine-tuned for $k$ shots in rare tissues. The same hyperparameters are used in training. The model was finally tested with AUPRC and AUROC.

GPT-2 and GPT-3 are directly fine-tuned on rare tissues with pre-trained parameters from regular GPT-2 and GPT-3 Ada. For each $k$ shot in each tissue, GPT-2 is fine-tuned for 4 epochs using a learning rate of 5e-5 and a weight decay of 0.01. The hyperparameters of GPT-3 are adjusted by OpenAI API based on the data size. The model was also fine-tuned for 4 epochs. GPT-2 and GPT-3 fine-tuned models were finally tested with AUPRC and AUROC.

\acks{XJ is CPRIT Scholar in Cancer Research (RR180012), and he was supported in part by Christopher Sarofim Family Professorship, UT Stars award, UTHealth startup, the National Institute of Health (NIH) under award number R01AG066749, R01AG066749-03S1, R01LM013712 and U01TR002062, and the National Science Foundation (NSF) \#2124789. This work is supported by NSF AI Center Grant (NSF 2019844), NSF-CSIRO (NSF 2303038), NIH Bridge2AI (OTA-21-008), and Bill \& Lewis Suit Professorship at the University of Texas at Austin. YK was supported in part by the NIH under award number R01AG066749-03S1 and CPRIT RR180012.}

\bibliography{DrugComb}

\begin{thebibliography}{51}
\providecommand{\natexlab}[1]{#1}
\providecommand{\url}[1]{\texttt{#1}}
\expandafter\ifx\csname urlstyle\endcsname\relax
  \providecommand{\doi}[1]{doi: #1}\else
  \providecommand{\doi}{doi: \begingroup \urlstyle{rm}\Url}\fi

\bibitem[Borthakur et~al.(2009)Borthakur, Faderl, Ravandi, Padmanabhan, Stock,
  Wu, Li, Curt, Tallman, and Minden]{Borthakur2009-wg}
G~Borthakur, S~Faderl, F~Ravandi, S~Padmanabhan, W~Stock, K~Wu, J~Li, G~Curt,
  M~Tallman, and M~Minden.
\newblock Clinical, pharmacokinetic ({PK)}, and pharmacodynamic findings from a
  phase {I} trial of an eg5 inhibitor ({AZD4877}) in patients with refractory
  acute myeloid leukemia ({AML}).
\newblock \emph{J. Clin. Orthod.}, 27\penalty0 (15\_suppl):\penalty0
  3580--3580, May 2009.

\bibitem[Brown et~al.(2020)Brown, Mann, Ryder, Subbiah, Kaplan, Dhariwal,
  Neelakantan, Shyam, Sastry, Askell, Agarwal, Herbert-Voss, Krueger, Henighan,
  Child, Ramesh, Ziegler, Wu, Winter, Hesse, Chen, Sigler, Litwin, Gray, Chess,
  Clark, Berner, McCandlish, Radford, Sutskever, and
  Amodei]{brown_language_2020}
Tom~B. Brown, Benjamin Mann, Nick Ryder, Melanie Subbiah, Jared Kaplan,
  Prafulla Dhariwal, Arvind Neelakantan, Pranav Shyam, Girish Sastry, Amanda
  Askell, Sandhini Agarwal, Ariel Herbert-Voss, Gretchen Krueger, Tom Henighan,
  Rewon Child, Aditya Ramesh, Daniel~M. Ziegler, Jeffrey Wu, Clemens Winter,
  Christopher Hesse, Mark Chen, Eric Sigler, Mateusz Litwin, Scott Gray,
  Benjamin Chess, Jack Clark, Christopher Berner, Sam McCandlish, Alec Radford,
  Ilya Sutskever, and Dario Amodei.
\newblock Language {Models} are {Few}-{Shot} {Learners}, July 2020.
\newblock URL \url{http://arxiv.org/abs/2005.14165}.
\newblock arXiv:2005.14165 [cs].

\bibitem[Celebi et~al.(2019)Celebi, Bear Don't~Walk, Movva, Alpsoy, and
  Dumontier]{Celebi2019-nl}
Remzi Celebi, Oliver Bear Don't~Walk, 4th, Rajiv Movva, Semih Alpsoy, and
  Michel Dumontier.
\newblock In-silico prediction of synergistic {Anti-Cancer} drug combinations
  using multi-omics data.
\newblock \emph{Sci. Rep.}, 9\penalty0 (1):\penalty0 8949, June 2019.

\bibitem[Cervantes-Gomez et~al.(2019)Cervantes-Gomez, Stellrecht, Ayres,
  Keating, Wierda, and Gandhi]{Cervantes-Gomez2019-de}
Fabiola Cervantes-Gomez, Christine~M Stellrecht, Mary~L Ayres, Michael~J
  Keating, William~G Wierda, and Varsha Gandhi.
\newblock {PIM} kinase inhibitor, {AZD1208}, inhibits protein translation and
  induces autophagy in primary chronic lymphocytic leukemia cells.
\newblock \emph{Oncotarget}, 10\penalty0 (29):\penalty0 2793--2809, April 2019.

\bibitem[Chen and Li(2018)]{chen_drugcom_2018}
Huiyuan Chen and Jing Li.
\newblock {DrugCom}: {Synergistic} {Discovery} of {Drug} {Combinations} {Using}
  {Tensor} {Decomposition}.
\newblock In \emph{2018 {IEEE} {International} {Conference} on {Data} {Mining}
  ({ICDM})}, pages 899--904, November 2018.
\newblock \doi{10.1109/ICDM.2018.00108}.
\newblock ISSN: 2374-8486.

\bibitem[Chen and Guestrin(2016)]{Chen2016-bv}
Tianqi Chen and Carlos Guestrin.
\newblock {XGBoost}: A scalable tree boosting system.
\newblock March 2016.

\bibitem[Cheng et~al.(2019)Cheng, Kov{\'a}cs, and Barab{\'a}si]{Cheng2019-dp}
Feixiong Cheng, Istv{\'a}n~A Kov{\'a}cs, and Albert-L{\'a}szl{\'o}
  Barab{\'a}si.
\newblock Network-based prediction of drug combinations.
\newblock \emph{Nat. Commun.}, 10\penalty0 (1):\penalty0 1197, March 2019.

\bibitem[Cortes et~al.(2018)Cortes, Tamura, DeAngelo, de~Bono, Lorente, Minden,
  Uy, Kantarjian, Chen, Gandhi, Godin, Keating, McEachern, Vishwanathan, Pease,
  and Dean]{Cortes2018-tx}
Jorge Cortes, Kenji Tamura, Daniel~J DeAngelo, Johann de~Bono, David Lorente,
  Mark Minden, Geoffrey~L Uy, Hagop Kantarjian, Lisa~S Chen, Varsha Gandhi,
  Robert Godin, Karen Keating, Kristen McEachern, Karthick Vishwanathan,
  Janet~Elizabeth Pease, and Emma Dean.
\newblock Phase {I} studies of {AZD1208}, a proviral integration moloney virus
  kinase inhibitor in solid and haematological cancers.
\newblock \emph{Br. J. Cancer}, 118\penalty0 (11):\penalty0 1425--1433, May
  2018.

\bibitem[Finn et~al.(2017)Finn, Abbeel, and Levine]{finn_model-agnostic_2017}
Chelsea Finn, Pieter Abbeel, and Sergey Levine.
\newblock Model-{Agnostic} {Meta}-{Learning} for {Fast} {Adaptation} of {Deep}
  {Networks}, July 2017.
\newblock URL \url{http://arxiv.org/abs/1703.03400}.
\newblock arXiv:1703.03400 [cs].

\bibitem[Gao et~al.(2023)Gao, Gao, Fan, Zhu, Wei, Zhou, Chuai, Chen, Zhang, and
  Liu]{Gao2023-se}
Yicheng Gao, Yuli Gao, Yuxiao Fan, Chengyu Zhu, Zhiting Wei, Chi Zhou, Guohui
  Chuai, Qinchang Chen, He~Zhang, and Qi~Liu.
\newblock {Pan-Peptide} meta learning for t-cell receptor--antigen binding
  recognition.
\newblock \emph{Nature Machine Intelligence}, 5\penalty0 (3):\penalty0
  236--249, March 2023.

\bibitem[Greco et~al.(1995)Greco, Bravo, and Parsons]{greco_search_1995}
W~R Greco, G~Bravo, and J~C Parsons.
\newblock The search for synergy: a critical review from a response surface
  perspective.
\newblock \emph{Pharmacol. Rev.}, 47\penalty0 (2):\penalty0 331--385, June
  1995.

\bibitem[Guo et~al.(2011)Guo, Dial, Shi, Branham, Liu, Fang, Green, Deng,
  Kaput, and Ning]{guo_similarities_2011}
Lei Guo, Stacey Dial, Leming Shi, William Branham, Jie Liu, Jia-Long Fang,
  Bridgett Green, Helen Deng, Jim Kaput, and Baitang Ning.
\newblock Similarities and {Differences} in the {Expression} of
  {Drug}-{Metabolizing} {Enzymes} between {Human} {Hepatic} {Cell} {Lines} and
  {Primary} {Human} {Hepatocytes}.
\newblock \emph{Drug Metabolism and Disposition}, 39\penalty0 (3):\penalty0
  528--538, March 2011.
\newblock ISSN 0090-9556.
\newblock \doi{10.1124/dmd.110.035873}.
\newblock URL \url{https://www.ncbi.nlm.nih.gov/pmc/articles/PMC3061558/}.

\bibitem[Guo et~al.(2010)Guo, Mao, Chen, Huang, Jin, Xu, and Qiu]{Guo2010-ly}
Shengjie Guo, Xiaopeng Mao, Junxing Chen, Bin Huang, Chu Jin, Zhenbo Xu, and
  Shaopeng Qiu.
\newblock Overexpression of pim-1 in bladder cancer.
\newblock \emph{J. Exp. Clin. Cancer Res.}, 29\penalty0 (1):\penalty0 161,
  December 2010.

\bibitem[He et~al.(2018)He, Kulesskiy, Saarela, Turunen, Wennerberg,
  Aittokallio, and Tang]{he_methods_2018}
Liye He, Evgeny Kulesskiy, Jani Saarela, Laura Turunen, Krister Wennerberg,
  Tero Aittokallio, and Jing Tang.
\newblock Methods for {High}-throughput {Drug} {Combination} {Screening} and
  {Synergy} {Scoring}.
\newblock In Louise von Stechow, editor, \emph{Cancer {Systems} {Biology}:
  {Methods} and {Protocols}}, Methods in {Molecular} {Biology}, pages 351--398.
  Springer, New York, NY, 2018.
\newblock ISBN 978-1-4939-7493-1.
\newblock \doi{10.1007/978-1-4939-7493-1_17}.
\newblock URL \url{https://doi.org/10.1007/978-1-4939-7493-1_17}.

\bibitem[Hegselmann et~al.(2023)Hegselmann, Buendia, Lang, Agrawal, Jiang, and
  Sontag]{hegselmann_tabllm_2023}
Stefan Hegselmann, Alejandro Buendia, Hunter Lang, Monica Agrawal, Xiaoyi
  Jiang, and David Sontag.
\newblock {TabLLM}: {Few}-shot {Classification} of {Tabular} {Data} with
  {Large} {Language} {Models}, March 2023.
\newblock URL \url{http://arxiv.org/abs/2210.10723}.
\newblock arXiv:2210.10723 [cs].

\bibitem[Hosseini and Zhou(2023)]{hosseini_ccsynergy_2023}
Sayed-Rzgar Hosseini and Xiaobo Zhou.
\newblock {CCSynergy}: an integrative deep-learning framework enabling
  context-aware prediction of anti-cancer drug synergy.
\newblock \emph{Briefings in Bioinformatics}, 24\penalty0 (1):\penalty0
  bbac588, January 2023.
\newblock ISSN 1477-4054.
\newblock \doi{10.1093/bib/bbac588}.
\newblock URL \url{https://doi.org/10.1093/bib/bbac588}.

\bibitem[Huang et~al.(2020)Huang, Khetan, Cvitkovic, and
  Karnin]{huang_tabtransformer_2020}
Xin Huang, Ashish Khetan, Milan Cvitkovic, and Zohar Karnin.
\newblock {TabTransformer}: {Tabular} {Data} {Modeling} {Using} {Contextual}
  {Embeddings}, December 2020.
\newblock URL \url{http://arxiv.org/abs/2012.06678}.
\newblock arXiv:2012.06678 [cs].

\bibitem[Ianevski et~al.(2017)Ianevski, He, Aittokallio, and
  Tang]{ianevski_synergyfinder_2017}
Aleksandr Ianevski, Liye He, Tero Aittokallio, and Jing Tang.
\newblock {SynergyFinder}: a web application for analyzing drug combination
  dose-response matrix data.
\newblock \emph{Bioinformatics}, 33\penalty0 (15):\penalty0 2413--2415, August
  2017.

\bibitem[Jones et~al.(2013)Jones, Vuky, Elliott, Mead, Arranz, Chester,
  Chowdhury, Dudek, M{\"u}ller-Mattheis, Grimm, Gschwend, W{\"u}lfing, Albers,
  Li, Osmukhina, Skolnik, and Hudes]{Jones2013-zl}
Robert Jones, Jacqueline Vuky, Tony Elliott, Graham Mead, Jos{\'e}~Angel
  Arranz, John Chester, Simon Chowdhury, Arkadiusz~Z Dudek, Volker
  M{\"u}ller-Mattheis, Marc-Oliver Grimm, J{\"u}rgen~E Gschwend, Christian
  W{\"u}lfing, Peter Albers, Jianguo Li, Anna Osmukhina, Jeffrey Skolnik, and
  Gary Hudes.
\newblock Phase {II} study to assess the efficacy, safety and tolerability of
  the mitotic spindle kinesin inhibitor {AZD4877} in patients with recurrent
  advanced urothelial cancer.
\newblock \emph{Invest. New Drugs}, 31\penalty0 (4):\penalty0 1001--1007,
  August 2013.

\bibitem[Kim et~al.(2021)Kim, Zheng, Tang, Jim~Zheng, Li, and
  Jiang]{kim_anticancer_2021}
Yejin Kim, Shuyu Zheng, Jing Tang, Wenjin Jim~Zheng, Zhao Li, and Xiaoqian
  Jiang.
\newblock Anticancer drug synergy prediction in understudied tissues using
  transfer learning.
\newblock \emph{Journal of the American Medical Informatics Association},
  28\penalty0 (1):\penalty0 42--51, January 2021.
\newblock ISSN 1527-974X.
\newblock \doi{10.1093/jamia/ocaa212}.
\newblock URL \url{https://doi.org/10.1093/jamia/ocaa212}.

\bibitem[Kuru et~al.(2022)Kuru, Tastan, and Cicek]{kuru_matchmaker_2022}
Halil~Ibrahim Kuru, Oznur Tastan, and A.~Ercument Cicek.
\newblock {MatchMaker}: {A} {Deep} {Learning} {Framework} for {Drug} {Synergy}
  {Prediction}.
\newblock \emph{IEEE/ACM transactions on computational biology and
  bioinformatics}, 19\penalty0 (4):\penalty0 2334--2344, 2022.
\newblock ISSN 1557-9964.
\newblock \doi{10.1109/TCBB.2021.3086702}.

\bibitem[Li et~al.(2018)Li, Li, Quang, and Guan]{li_network_2018}
Hongyang Li, Tingyang Li, Daniel Quang, and Yuanfang Guan.
\newblock Network {Propagation} {Predicts} {Drug} {Synergy} in {Cancers}.
\newblock \emph{Cancer Research}, 78\penalty0 (18):\penalty0 5446--5457,
  September 2018.
\newblock ISSN 0008-5472.
\newblock \doi{10.1158/0008-5472.CAN-18-0740}.
\newblock URL \url{https://doi.org/10.1158/0008-5472.CAN-18-0740}.

\bibitem[Li et~al.(2020)Li, Li, Suhara, Doan, and Tan]{li_deep_2020}
Yuliang Li, Jinfeng Li, Yoshihiko Suhara, AnHai Doan, and Wang-Chiew Tan.
\newblock Deep {Entity} {Matching} with {Pre}-{Trained} {Language} {Models}.
\newblock \emph{Proceedings of the VLDB Endowment}, 14\penalty0 (1):\penalty0
  50--60, September 2020.
\newblock ISSN 2150-8097.
\newblock \doi{10.14778/3421424.3421431}.
\newblock URL \url{http://arxiv.org/abs/2004.00584}.
\newblock arXiv:2004.00584 [cs].

\bibitem[Lin et~al.(2020)Lin, Patil, Gao, and Qian]{lin_bone_2020}
Xiao Lin, Suryaji Patil, Yong-Guang Gao, and Airong Qian.
\newblock The {Bone} {Extracellular} {Matrix} in {Bone} {Formation} and
  {Regeneration}.
\newblock \emph{Frontiers in Pharmacology}, 11, 2020.
\newblock \doi{10.3389/fphar.2020.00757}.
\newblock URL \url{https://www.ncbi.nlm.nih.gov/pmc/articles/PMC7264100/}.
\newblock Publisher: Frontiers Media SA.

\bibitem[Liu and Xie(2021)]{liu_transynergy_2021}
Qiao Liu and Lei Xie.
\newblock {TranSynergy}: {Mechanism}-driven interpretable deep neural network
  for the synergistic prediction and pathway deconvolution of drug
  combinations.
\newblock \emph{PLOS Computational Biology}, 17\penalty0 (2):\penalty0
  e1008653, February 2021.
\newblock ISSN 1553-7358.
\newblock \doi{10.1371/journal.pcbi.1008653}.
\newblock URL
  \url{https://journals.plos.org/ploscompbiol/article?id=10.1371/journal.pcbi.1008653}.
\newblock Publisher: Public Library of Science.

\bibitem[Liu et~al.(2022)Liu, Song, Liu, Li, Zhou, and
  Zhang]{10.1093/bioinformatics/btac579}
Xuan Liu, Congzhi Song, Shichao Liu, Menglu Li, Xionghui Zhou, and Wen Zhang.
\newblock {Multi-way relation-enhanced hypergraph representation learning for
  anti-cancer drug synergy prediction}.
\newblock \emph{Bioinformatics}, 38\penalty0 (20):\penalty0 4782--4789, 08
  2022.
\newblock ISSN 1367-4803.
\newblock \doi{10.1093/bioinformatics/btac579}.
\newblock URL \url{https://doi.org/10.1093/bioinformatics/btac579}.

\bibitem[Madani et~al.(2023)Madani, Krause, Greene, Subramanian, Mohr, Holton,
  Olmos, Xiong, Sun, Socher, Fraser, and Naik]{Madani2023-sp}
Ali Madani, Ben Krause, Eric~R Greene, Subu Subramanian, Benjamin~P Mohr,
  James~M Holton, Jose~Luis Olmos, Jr, Caiming Xiong, Zachary~Z Sun, Richard
  Socher, James~S Fraser, and Nikhil Naik.
\newblock Large language models generate functional protein sequences across
  diverse families.
\newblock \emph{Nat. Biotechnol.}, January 2023.

\bibitem[Mitchell and Krakauer(2023)]{Mitchell2023-cs}
Melanie Mitchell and David~C Krakauer.
\newblock The debate over understanding in {AI's} large language models.
\newblock \emph{Proceedings of the National Academy of Sciences}, 120\penalty0
  (13):\penalty0 e2215907120, 2023.

\bibitem[Moor et~al.(2023)Moor, Banerjee, Abad, Krumholz, Leskovec, Topol, and
  Rajpurkar]{Moor2023-dp}
Michael Moor, Oishi Banerjee, Zahra Shakeri~Hossein Abad, Harlan~M Krumholz,
  Jure Leskovec, Eric~J Topol, and Pranav Rajpurkar.
\newblock Foundation models for generalist medical artificial intelligence.
\newblock \emph{Nature}, 616\penalty0 (7956):\penalty0 259--265, April 2023.

\bibitem[Nam et~al.(2023)Nam, Tack, Lee, Lee, and Shin]{nam_stunt_2023}
Jaehyun Nam, Jihoon Tack, Kyungmin Lee, Hankook Lee, and Jinwoo Shin.
\newblock Stunt: Few-shot tabular learning with self-generated tasks from
  unlabeled tables.
\newblock 2023.

\bibitem[Narayan et~al.(2022)Narayan, Chami, Orr, Arora, and
  Ré]{narayan_can_2022}
Avanika Narayan, Ines Chami, Laurel Orr, Simran Arora, and Christopher Ré.
\newblock Can {Foundation} {Models} {Wrangle} {Your} {Data}?, December 2022.
\newblock URL \url{http://arxiv.org/abs/2205.09911}.
\newblock arXiv:2205.09911 [cs].

\bibitem[{National Cancer Institute}(2011)]{noauthor_2011-op}
{National Cancer Institute}.
\newblock {NCI} drug dictionary.
\newblock
  \url{https://www.cancer.gov/publications/dictionaries/cancer-drug/def/pan-pim-kinase-inhibitor-azd1208},
  February 2011.
\newblock Accessed: 2023-4-13.

\bibitem[{NVIDIA}(2023)]{nvidia_bionemo}
{NVIDIA}.
\newblock {NVIDIA} {BioNeMo} cloud service: An end-to-end {AI-powered} drug
  discovery pipelines.
\newblock \url{https://www.nvidia.com/en-us/gpu-cloud/bionemo/}, 2023.
\newblock Accessed: 2023-4-14.

\bibitem[OpenAI(2021)]{openai_finetuning}
OpenAI.
\newblock Fine-tuning - openai api, 2021.
\newblock URL \url{https://platform.openai.com/docs/guides/fine-tuning}.

\bibitem[OpenAI(2022)]{openai_chatgpt}
OpenAI.
\newblock Introducing chatgpt, 2022.
\newblock URL \url{https://openai.com/blog/chatgpt}.

\bibitem[{OpenAI}(2023)]{OpenAI2023-ce}
{OpenAI}.
\newblock {GPT-4} technical report.
\newblock March 2023.

\bibitem[Preuer et~al.(2018)Preuer, Lewis, Hochreiter, Bender, Bulusu, and
  Klambauer]{preuer_deepsynergy_2018}
Kristina Preuer, Richard P~I Lewis, Sepp Hochreiter, Andreas Bender, Krishna~C
  Bulusu, and Günter Klambauer.
\newblock {DeepSynergy}: predicting anti-cancer drug synergy with {Deep}
  {Learning}.
\newblock \emph{Bioinformatics}, 34\penalty0 (9):\penalty0 1538--1546, May
  2018.

\bibitem[Radford et~al.(2018)Radford, Wu, Child, Luan, Amodei, and
  Sutskever]{radford_language_nodate}
Alec Radford, Jeffrey Wu, Rewon Child, David Luan, Dario Amodei, and Ilya
  Sutskever.
\newblock Language {Models} are {Unsupervised} {Multitask} {Learners}.
\newblock 2018.

\bibitem[Schramowski et~al.(2022)Schramowski, Turan, Andersen, Rothkopf, and
  Kersting]{Schramowski2022-yg}
Patrick Schramowski, Cigdem Turan, Nico Andersen, Constantin~A Rothkopf, and
  Kristian Kersting.
\newblock Large pre-trained language models contain human-like biases of what
  is right and wrong to do.
\newblock \emph{Nature Machine Intelligence}, 4\penalty0 (3):\penalty0
  258--268, March 2022.

\bibitem[Sidorov et~al.(2019)Sidorov, Naulaerts, Ariey-Bonnet, Pasquier, and
  Ballester]{Sidorov2019-np}
Pavel Sidorov, Stefan Naulaerts, J{\'e}r{\'e}my Ariey-Bonnet, Eddy Pasquier,
  and Pedro~J Ballester.
\newblock Predicting synergism of cancer drug combinations using {NCI-ALMANAC}
  data.
\newblock \emph{Front Chem}, 7:\penalty0 509, July 2019.

\bibitem[Sun et~al.(2020)Sun, Huang, Jiang, and
  Hu]{10.1093/bioinformatics/btaa287}
Zexuan Sun, Shujun Huang, Peiran Jiang, and Pingzhao Hu.
\newblock {DTF: Deep Tensor Factorization for predicting anticancer drug
  synergy}.
\newblock \emph{Bioinformatics}, 36\penalty0 (16):\penalty0 4483--4489, 05
  2020.
\newblock ISSN 1367-4803.
\newblock \doi{10.1093/bioinformatics/btaa287}.
\newblock URL \url{https://doi.org/10.1093/bioinformatics/btaa287}.

\bibitem[Suphavilai et~al.(2018)Suphavilai, Bertrand, and
  Nagarajan]{10.1093/bioinformatics/bty452}
Chayaporn Suphavilai, Denis Bertrand, and Niranjan Nagarajan.
\newblock {Predicting Cancer Drug Response using a Recommender System}.
\newblock \emph{Bioinformatics}, 34\penalty0 (22):\penalty0 3907--3914, 06
  2018.
\newblock ISSN 1367-4803.
\newblock \doi{10.1093/bioinformatics/bty452}.
\newblock URL \url{https://doi.org/10.1093/bioinformatics/bty452}.

\bibitem[Tang and Gottlieb(2022)]{Tang2022-wn}
Yi-Ching Tang and Assaf Gottlieb.
\newblock {SynPathy}: Predicting drug synergy through {Drug-Associated}
  pathways using deep learning.
\newblock \emph{Mol. Cancer Res.}, 20\penalty0 (5):\penalty0 762--769, May
  2022.

\bibitem[Veit et~al.(2017)Veit, Alldrin, Chechik, Krasin, Gupta, and
  Belongie]{veit_learning_2017}
Andreas Veit, Neil Alldrin, Gal Chechik, Ivan Krasin, Abhinav Gupta, and Serge
  Belongie.
\newblock Learning {From} {Noisy} {Large}-{Scale} {Datasets} {With} {Minimal}
  {Supervision}, April 2017.
\newblock URL \url{http://arxiv.org/abs/1701.01619}.
\newblock arXiv:1701.01619 [cs].

\bibitem[Wang et~al.(2023)Wang, He, Yu, and Xu]{Wang2023-iw}
Duolin Wang, Fei He, Yang Yu, and Dong Xu.
\newblock Meta-learning for {T} cell receptor binding specificity and beyond.
\newblock \emph{Nature Machine Intelligence}, pages 1--3, March 2023.

\bibitem[Wertheimer and Hariharan(2019)]{wertheimer_few-shot_2019}
Davis Wertheimer and Bharath Hariharan.
\newblock Few-{Shot} {Learning} with {Localization} in {Realistic} {Settings},
  July 2019.
\newblock URL \url{http://arxiv.org/abs/1904.08502}.
\newblock arXiv:1904.08502 [cs, stat].

\bibitem[Wolf et~al.(2020)Wolf, Debut, Sanh, Chaumond, Delangue, Moi, Cistac,
  Rault, Louf, Funtowicz, Davison, Shleifer, von Platen, Ma, Jernite, Plu, Xu,
  Scao, Gugger, Drame, Lhoest, and Rush]{wolf2020huggingfaces}
Thomas Wolf, Lysandre Debut, Victor Sanh, Julien Chaumond, Clement Delangue,
  Anthony Moi, Pierric Cistac, Tim Rault, Rémi Louf, Morgan Funtowicz, Joe
  Davison, Sam Shleifer, Patrick von Platen, Clara Ma, Yacine Jernite, Julien
  Plu, Canwen Xu, Teven~Le Scao, Sylvain Gugger, Mariama Drame, Quentin Lhoest,
  and Alexander~M. Rush.
\newblock Huggingface's transformers: State-of-the-art natural language
  processing, 2020.

\bibitem[Yadav et~al.(2015)Yadav, Wennerberg, Aittokallio, and
  Tang]{yadav_searching_2015}
Bhagwan Yadav, Krister Wennerberg, Tero Aittokallio, and Jing Tang.
\newblock Searching for {Drug} {Synergy} in {Complex} {Dose}-{Response}
  {Landscapes} {Using} an {Interaction} {Potency} {Model}.
\newblock \emph{Comput. Struct. Biotechnol. J.}, 13:\penalty0 504--513,
  September 2015.

\bibitem[Yang et~al.(2022)Yang, Li, Lim, Pan, Van~Truong, Wang, Li, Li, Xiao,
  Ding, Chen, Liu, Xie, Alvarado, Wang, and Chen]{Yang2022-ne}
Haitao Yang, Jiali Li, Kai~Zhuo Lim, Chuanji Pan, Tien Van~Truong, Qian Wang,
  Kerui Li, Shuo Li, Xiao Xiao, Meng Ding, Tianle Chen, Xiaoli Liu, Qian Xie,
  Pablo Valdivia~y Alvarado, Xiaonan Wang, and Po-Yen Chen.
\newblock Automatic strain sensor design via active learning and data
  augmentation for soft machines.
\newblock \emph{Nature Machine Intelligence}, 4\penalty0 (1):\penalty0 84--94,
  January 2022.

\bibitem[Zagidullin et~al.(2019)Zagidullin, Aldahdooh, Zheng, Wang, Wang, Saad,
  Malyutina, Jafari, Tanoli, Pessia, and Tang]{zagidullin_drugcomb_2019}
Bulat Zagidullin, Jehad Aldahdooh, Shuyu Zheng, Wenyu Wang, Yinyin Wang, Joseph
  Saad, Alina Malyutina, Mohieddin Jafari, Ziaurrehman Tanoli, Alberto Pessia,
  and Jing Tang.
\newblock {DrugComb}: an integrative cancer drug combination data portal.
\newblock \emph{Nucleic Acids Res.}, 47\penalty0 (W1):\penalty0 W43--W51, July
  2019.

\bibitem[Zheng et~al.(2021)Zheng, Aldahdooh, Shadbahr, Wang, Aldahdooh, Bao,
  Wang, and Tang]{10.1093/nar/gkab438}
Shuyu Zheng, Jehad Aldahdooh, Tolou Shadbahr, Yinyin Wang, Dalal Aldahdooh, Jie
  Bao, Wenyu Wang, and Jing Tang.
\newblock {DrugComb update: a more comprehensive drug sensitivity data
  repository and analysis portal}.
\newblock \emph{Nucleic Acids Research}, 49\penalty0 (W1):\penalty0 W174--W184,
  06 2021.
\newblock ISSN 0305-1048.
\newblock \doi{10.1093/nar/gkab438}.
\newblock URL \url{https://doi.org/10.1093/nar/gkab438}.

\end{thebibliography}



\end{document}